\documentclass[a4paper]{article}

\usepackage{INTERSPEECH2021}
\usepackage{multirow}
\usepackage{float}

\title{TransfoRNN: Capturing the Sequential Information in Self-Attention Representations for Language Modeling}
\name{Tze Yuang Chong, Xuyang Wang, Lin Yang, Junjie Wang}
\address{AI Lab, Lenovo Research, Beijing, China}
\email{\{tchong,wangxy60,yanglin13,wangjj9\}@lenovo.com}

\begin{document}

\maketitle
\begin{abstract}
In this paper, we describe the use of recurrent neural networks to capture sequential information from the self-attention representations to improve the Transformers. Although self-attention mechanism provides a means to exploit long context, the sequential information, i.e. the arrangement of tokens, is not explicitly captured. We propose to cascade the recurrent neural networks to the Transformers, which referred to as the TransfoRNN model, to capture the sequential information. We found that the TransfoRNN models which consists of only shallow Transformers stack is suffice to give comparable, if not better, performance than a deeper Transformer model. Evaluated on the Penn Treebank and WikiText-2 corpora, the proposed TransfoRNN model has shown lower model perplexities with fewer number of model parameters. On the Penn Treebank corpus, the model perplexities were reduced up to 5.5\% with the model size reduced up to 10.5\%. On the WikiText-2 corpus, the model perplexity was reduced up to 2.2\% with a 27.7\% smaller model. Also, the TransfoRNN model was applied on the LibriSpeech speech recognition task and has shown comparable results with the Transformer models.

\end{abstract}
\noindent\textbf{Index Terms}: transformers, language model, speech recognition

\section{Introduction}

The Transformers have shown to outperform the vanilla recurrent neural networks in many natural language processing tasks \cite{Vaswani2017, Peters2018, Liu2019, Devlin2019, Fedus2021}. One of the key factors of such improvement is the self-attention mechanism which allows contexts to be processed based on relation among tokens \cite{Bahdanau2015}. Specifically, each token is attended to all other tokens in the context, thus offers an alternative to the recurrence mechanism to capture the distant information. Also, as the recurrent neural networks have shown to focus on only the nearest 50 tokens \cite{Khandelwal2018}, the self-attention mechanism provides a better solution to exploit longer contexts.



Applying the Transformers in a causal manner is crucial for language modeling, particularly predicting next word given the preceding context \cite{Peters2018, Irie2019, Al-Rfou2019, Irie2020}. Typically, a window is slid through a long sequence of tokens, in which the self-attention representations are computed and used to estimate probabilities. However, as self-attention itself has no regards on how tokens are arranged, the sequential information is usually compensated by encoding the embedding vectors based on positions \cite{Vaswani2017}.  

In this paper, we investigate if such sequential information can be captured in a more explicit manner. That is, we attempt to preserve the arrangement of the self-attention representations and use such information to improve the language models. The implementation involves cascading a standard recurrent neural network (RNN) to the Transformer layers, hence we refer this architecture to as TransfoRNN. Through a series of evaluations, we show empirically that the TransfoRNN models outperform the vanilla Transformers models, in terms of model perplexities and speech recognition accuracy. Furthermore, we discovered that with the inclusion of the sequential information into the model, a TransfoRNN model with shallow Transformer layers, e.g. two layers, is suffice to give comparable performance, if not better, to a deeper Transformer network. Also, we highlight that the TransfoRNN models possess faster inference time.

Next section provides some background information about language modeling. In Section  and we will highlight the Transformers and some related works in language modeling. In Section \ref{sec:problem}, we highlight the shortcoming of using the Transformers for language modeling. Next in Section \ref{sec:transfornn}, we depict our proposed TransfoRNN model. Section \ref{sec:experiments} presents the experiments setup and results particularly the comparison of the TransforRNN model with the standard Transformers. Finally, Section \ref{sec:conclusion} concludes this paper and suggests future works.

\section{Background}
\label{sec:background}


The function of a language model (LM) is to estimate the probability distribution of next word given the history context, i.e. $P(w_i|w_1^{i-1})$. Traditionally, LMs are estimated from the smoothed $n$-gram counts \cite{Jelinek1985, Katz1987, Witten1991, Kneser1995, Chen1996}. In the past two decades, neural networks have been extensively used for language modeling due to their better smoothing capability, particularly when using longer history context, i.e. $|w_1^{i-1}| \gg 3,4$. Some commonly used architectures are the feedforward \cite{Bengio2001,Schwenk2007} and recurrent neural networks \cite{Mikolov2010,Sundermeyer2012,Arisoy2015}, and recently, the Transformers \cite{Al-Rfou2019,Irie2019,Irie2020,Beck2020}. The idea of these approaches is to project discrete words into low dimensional space such that latent information about the context can be better captured. 

In general, neural network approach to language modeling can be depicted as follows.

\begin{equation}
\label{eq:nnlm}
P(w_i|w_1^{i-1}) \approx P_{\text{NN}} \big(w_i|f(w_1^{i-1})\big)
\end{equation}
where $f(\cdot)$ refers to projection applied on the history context.

The LMs are devised to capture as much knowledge from the language, especially the syntax, which governs how tokens, e.g. words, are arranged in sequence. In order to capture the sequential information, a feed-forward neural network (FNN) concatenates the embedding vectors of the tokens in the order as how they are arranged in the history context.
\begin{equation}
\label{eq:fnnlm}
P_{\text{FNN}} \big(w_i|<e_1, e_2, ..., e_{i-1}>\big)
\end{equation} where $e_1, e_2, ..., e_{i-1}$ denote the embedding vectors of tokens $w_1, w_2, ..., w_{i-1}$. 

A recurrent neural network (RNN), on the other hand, digests the history context recursively, i.e. token by token, to update the context vector in the network.
\begin{equation}
\label{eq:rnnlm}
P_{\text{RNN}} \big(w_i|c_1^{i-2},e_{i-1}\big)
\end{equation} where $c_1^{i-2}$ and $e_{i-1}$ refer to the embedding vectors of context and token, respectively. 

For the Transformers, next token is predicted based on the self-attention representation of the immediate preceding token, which is computed from its embedding vector, embellished by attending it to other tokens in the history context through multiple Transformer layers \cite{Al-Rfou2019,Irie2019,Irie2020,Beck2020}.
\begin{equation}
\label{eq:translm}
P_{\text{Transformer}} \big(w_i|z_{i-1}^{(N)}\big)
\end{equation} where $z_{i-1}^{(N)}$ denotes the self-attention representation of token $w_{i-1}$ computed by a stack of $N$ layers of Transformer.


\section{Problem Statement}
\label{sec:problem}

Self attention mechanism serves the core module in the Transformers. It allows distant information to be captured and incorporated into the local representation by attending each token to other tokens in the entire context \cite{Vaswani2017}. The computation begins by first mapping the embedding vector of a given token to query, key and value vectors.

\begin{equation*}
q_{t}, k_{t}, v_{t} = Qe_{t}, Ke_{t}, Ve_{t}
\end{equation*} where $e_{t}$ denotes the embedding vector corresponds to token $w_t$ and $Q$, $K$ and $V$ are the functions projecting $e_t$ to its respective query, key and value vectors. The self-attention of $w_t$ can be computed as the weighted sum of the value vectors, each is weighted by how similar its corresponding query to other keys. The similarity score is normalized by using a softmax function. 

\begin{equation}
\label{eq:self-attention}
h_t = \sum_{i<t} \frac{\text{exp}(q_t \cdot k_i)}{\sum_{j<t} \text{exp}(q_t \cdot k_j)} v_i
\end{equation} Note that token $w_t$ attends only to the preceding tokens, i.e. $w_1, w_2, ..., w_{t-1}$, which reflects to the causality property in the language. In the actual realization, multiple sets of $Q$, $K$ and $V$ will be applied and the resulted representations are concatenated. This is referred to as the multi-head attention mechanism \cite{Vaswani2017}. 

%

Since the attention mechanism does not take any positional information into account (see Eq.\ref{eq:self-attention}), in order to compensate such deficiency, the embedding vectors are usually pre-encoded based on their positions.

\begin{equation}
\label{eq:pe}
e_t=e_t^{\text{RAW}} + e_t^{\text{POS}}
\end{equation} where $e_t^{\text{RAW}}$ is the raw embedding vector obtained from linear transformation, while $e_t^{\text{POS}}$ is the positional encoder typically implemented as sinusoidal functions with various frequencies.

The positional encoder ensures that identical tokens located in different positions will be represented as different vectors. However, how such encoding scheme would reflect the sequential information, i.e. the arrangement of the tokens, is unclear. As pointed out also in other works \cite{Tang2018, Raganato2019}, the Transformers may not capture the syntatic information as well as the RNNs. For language modeling, particularly, as farther tokens would generally have weaker predictive power towards next word \cite{Chong2013}, the positional encoding does not reflect this phenomenon.



\section{TransfoRNN}
\label{sec:transfornn}

As discussed previously, the FNN and RNN models attempt to capture the sequential information either by concatenating the embedding vectors or presenting the vectors sequentially to the network. Although the Transformers does modify the raw embedding vectors based on positions (see Eq.\ref{eq:pe}), how sequential information can be captured is doubtful.


\subsection{Architecture}

\begin{figure}[H]
	\centering
	\includegraphics[scale=0.6]{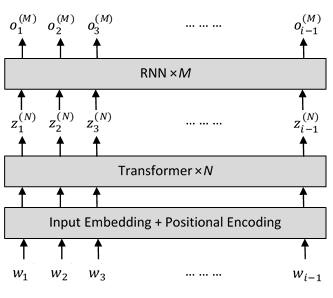}
	\caption{The architecture of the TransfoRNN model. The structure consists of $N$ layers of Transformers, cascaded by $M$ layers of RNNs.}
	\label{fig:transfornn}
\end{figure}

To capture the sequential information in a more explicit manner, we propose to cascade the RNNs to the Transformers, such that the the arrangement in the token sequence can be modeled, but at the self-attention level. The benefits, besides keeping the strength of the self-attention mechanism in the models, the sequential information can be preserved. We refer this proposed model to as TransfoRNN. The architecture is shown in Figure \ref{fig:transfornn}.

As compared to the Transformers which condition only on single representation (see Eq.\ref{eq:translm}), the TransfoRNN models make use of the entire representations in the context. Moreover, using the RNNs to process the representations of a long context allows the memory to be feasibly maintained during computation \cite{Irie2020}. 


\subsection{Implementation}

The Transformer layers in the TransfoRNN models are realized followed the original implementation \cite{Vaswani2017}. In the $l$-th layer, the self-attention component (see Eq.\ref{eq:self-attention}) is surrounded by a residual connection followed by layer normalization \cite{Ba2016}.

\begin{equation}
x_t^{(l)} = \text{LayerNorm}(W_0h_t^{(l)} + z_t^{(l-1)})
\end{equation} where $W_0$ denotes the linear function of the residual connection. Note that, at the first layer, $z_t^{0} = e_t$. Next, a feed-forward component is applied and surrounded by another residual connection and layer normalization.

\begin{equation}
\label{eq:trans_ff}
y_t^{(l)} = x_t^{(l)} + \text{Activation}(W_{1}x_t^{(l)})
\end{equation}
\begin{equation}
z_t^{(l)} = \text{LayerNorm}(y_t^{(l)})
\end{equation} where $W_1$ refers to the linear function in the feed-forward connection. 

At the $N$-th layer of the Transformers, a stack of $M$ layers of RNNs takes over the representation, i.e. $z_1^{(N)}, z_2^{(N)}, ..., z_{i-1}^{(N)}$, to produce outputs, $o_1^{(M)}, o_2^{(M)}, ..., o_{i-1}^{(M)}$. Probabilities can be obtained by sending the outputs through a softmax layer.

\section{Experiments}
\label{sec:experiments}

The performance of the proposed TransfoRNN model was evaluated in terms of perplexity (PPL) and speech recognition word error rate (WER). The TransfoRNN model was compared with the vanilla Transformer model to assess the performance gain contributed by the captured sequential information.

For the TransfoRNN models, the recurrent layers were implemented as the long short term memory (LSTM) neural networks \cite{Sundermeyer2012}. The Transformer models were trained followed the approach in \cite{Al-Rfou2019}, where during training, all outputs from the final layer, i.e. $z_1^{(N)}$, $z_2^{(N)}$, ..., $z_{i-1}^{(N)}$, were used to compute the loss, but only the last output, i.e. $z_{i-1}^{(N)}$, was used for inference. As will be shown later, this configuration would give lower model PPL as compared to using the entire outputs for prediction. In all models considered in the experiments, the input embedding was tied to the output embedding \cite{Press2017, Inan2017}. The models were built by using the PyTorch toolkit \cite{Paszke2019}.

\subsection{Perplexity Analysis}

The model PPLs were evaluated based on the Penn Treebank (PTB) and WikiText-2 \cite{Merity2017} corpora, each consists of 0.9M and 2.0M words in the training set. The vocabularies comprise 10K and 33K words, respectively. 

For both TransfoRNN and Transformer models, the number of heads in the Transformer layers was 8 and the dimension of the feed-forward component (see Eq.\ref{eq:trans_ff}) was 1024. The models were trained using the stochastic gradient descent (SGD) method and new-bob learning rate adjustment. The initial learning rate was commonly fixed as 0.1 and the batch size was 16.

\subsubsection{TransfoRNN vs. Transformer}

First of all, the TransfoRNN models were compared with the Transformer models. The number of LSTM layers in the TransfoRNN models was fixed at two, while the depth of the Transformer layers was increased from two to eight. For the Transformer models, the depths varied from two to sixteen. In this evaluation, the number of heads of the Transformer layers, in both types of models, was eight and the dimension of the feed-forward component is 1024 (see Eq.\ref{eq:trans_ff}). Both models were compared in the embedding dimensions of 512 and 1024. The PPLs are shown in Table \ref{tab:ppl_ptb_wiki2}.

\begin{table}[]
\caption{PPLs of the TransfoRNN models as compared to the Transformer models. On both corpora, TransfoRNN models with only two Transformer layers showed lowest PPLs. (\textit{d}: embedding dimension, \textit{N}: number of Transformer layers)}

	\centering
\begin{tabular}{|l|l|c|c|c|c|c|}
\hline
\multicolumn{2}{|l|}{\multirow{2}{*}{}}               & \multicolumn{1}{l|}{\multirow{2}{*}{\textit{N}}} & \multicolumn{2}{c|}{PTB} & \multicolumn{2}{c|}{WikiText-2} \\ \cline{4-7} 
\multicolumn{2}{|l|}{}                                 & \multicolumn{1}{l|}{}                   & PPL        & \#param     & PPL           & \#param         \\ \hline
& KN5                          & -                                       & 147.9      & -           & 231.0         & -               \\ \hline \hline 
\multirow{8}{*}{\rotatebox{90}{\textit{d}=512}}                        & \multirow{4}{*}{Transformer} & 2                                       & 103.0      & 9.3M        & 127.1         & 21.3M           \\  
                        &                              & 4                                       & 93.8       & 13.5M       & 108.5         & 25.5M           \\  
                        &                              & 8                                       & 88.5       & 22.0M       & 100.1         & 33.9M           \\  
                        &                              & 16                                      & 101.5      & 38.8M       & 100.0         & 50.7M           \\ \cline{2-7} 
                        & \multirow{3}{*}{TransfoRNN}  & 2                                       & \textbf{83.6}       & 18.7M       & \textbf{98.5}          & 42.5M           \\  
                        &                              & 4                                       & 86.1       & 22.9M       & 102.0         & 46.7M           \\  
                        &                              & 8                                       & 90.2       & 31.3M       & 109.1         & 55.1M           \\ \cline{2-7} 
                        & LSTM                         & -                                       & 99.0       & 9.3M        & 100.7         & 21.3M           \\ \hline \hline
\multirow{8}{*}{\rotatebox{90}{\textit{d}=1024}} & \multirow{4}{*}{Transformer} & 2                                       & 111.3      & 22.9M       & 130.9         & 46.7M           \\  
                        &                              & 4                                       & 98.2       & 35.5M       & 109.7         & 59.3M           \\  
                        &                              & 8                                       & 94.7       & 60.7M       & 98.6          & 84.5M           \\  
                        &                              & 16                                      & 100.3      & 111.1M      & 96.9          & 134.9M          \\ \cline{2-7} 
                        & \multirow{3}{*}{TransfoRNN}  & 2                                       & \textbf{83.2}       & 49.9M       & \textbf{94.8}          & 97.6M           \\  
                        &                              & 4                                       & 83.2       & 62.5M       & 98.4          & 110.2M          \\  
                        &                              & 8                                       & 89.9       & 87.7M       & 104.3         & 135.4M          \\ \cline{2-7} 
                        & LSTM                         & -                                       & 105.6      & 27.0M       & 114.4         & 50.9M           \\ \hline
\end{tabular}
 \label{tab:ppl_ptb_wiki2}
\end{table}

As shown by the results in Table \ref{tab:ppl_ptb_wiki2}, the TransfoRNN models outperformed the Transformer models in most of the settings. Particularly, the TransfoRNN models which comprise only two Transformer layers, i.e. $N=2$, outperformed a deeper Transformer models On the WikiText-2 corpus, for example, although deepening a Transformer model consistently reduced the model PPLs (from 130.9 to 96.9 as $N$ was increased from 2 to 8 in the setting of $d=1024$), a TransfoRNN model with merely 2 Transformer layers have shown a lower PPL, i.e. 94.8. Under the best setting of the Transformer models, the TransfoRNN models reduced the PPLs up to 5.5\% on the PTB corpus (from 88.5 to 83.6) and 2.2\% on the WikiText-2 corpus (from 96.9 to 94.8). 

More importantly, the TransfoRNN models demand fewer number of parameters in the models to achieve such comparable results. Particularly, on the PTB corpus, the model size was reduced by 10.5\% (from 22.0M to 18.7M), while on the WikiText-2 corpus, the model size was reduced by 27.7\% (from 134.9M to 97.6M). Both results were measured under the best settings of the models.

We notices that for the PTB corpus, the PPL of the Transformer model increased when the model went deeper than 8 layers. This can be explained by the PTB corpus is a comparatively small dataset which would easily cause the model overfits.

The LSTM models were also evaluated in order to validate the performance of the TransfoRNN models. Such results indicate the performance of the TransfoRNN models when the Transformer layers were unplugged. As expected, the TransfoRNN models outperform the LSTM and Transformer models.

\subsubsection{Number of RNN layers}

Next, we assessed the optimal number of LSTM layers, i.e. $M$, in the TransfoRNN models. By varying the number of LSTM layers, 2 layers of LSTM are found to be optimum to the TransfoRNN models. The results are shown in Table \ref{tab:ppl_rnn_layer}.

\begin{table}[H]
\caption{PPLs of the TranfoRNN models with different numbers of RNN layers. }
\centering
\begin{tabular}{|l|c|c|c|c|c|}
\hline
\multirow{2}{*}{\textit{d}} & \multirow{2}{*}{\textit{M}} & \multicolumn{2}{c|}{PTB} & \multicolumn{2}{c|}{WikiText-2} \\ \cline{3-6} 
                   &                    & PPL       & \#param      & PPL           & \#param         \\ \hline
512                & 1                  & 85.6      & 16.6M        & 100.3         & 40.4M           \\ 
                   & 2                  & \textbf{83.5}      & 18.7M        & \textbf{98.5}          & 42.5M           \\ 
                   & 3                  & 89.0      & 20.8M        & 106.5         & 44.6M           \\ \hline
1024               & 1                  & 89.6      & 41.5M        & 97.8          & 89.2M           \\ 
                   & 2                  & \textbf{83.2}      & 49.9M        & \textbf{94.8}          & 97.6M           \\ 
                   & 3                  & 88.9      & 58.3M        & 98.3          & 106.0M          \\ \hline
\end{tabular}
\label{tab:ppl_rnn_layer}
\end{table}

We noticed that the model PPLs were sensitive to the deeper LSTM stacks, particularly in the lower embedding dimension setting.

\subsubsection{With \& without positional encoding}

As the sequential information has been captured in the LSTM layers, we evaluated if the positional encoding (see Eq.\ref{eq:pe}) is still required in this scenario. We evaluated the TransfoRNN models with and without positional encoding. Under the situation where the positional encoder was removed, the input to the model is hence the raw embedding vectors. We compared the models of different embedding dimensions: 256, 512, 1024 and 2048. The results are in Table \ref{tab:ppl_position}.

\begin{table}[H]
\caption{PPLs of the TransfoRNN models with and without positional encoding under different dimensions of embedding.}
\centering
\begin{tabular}{|l|c|c|c|c|}
\hline
\multirow{2}{*}{} & \multicolumn{2}{c|}{PTB} & \multicolumn{2}{c|}{WikiText-2} \\ \cline{2-5} 
                  & with pos.  & w/o pos.    & with pos.      & w/o pos.       \\ \hline
256               & 93.1       & 94.0        & 114.9          & 107.8          \\ \hline
512               & 83.6       & 83.8        & 98.5           & 97.0           \\ \hline
1024              & 83.2       & \textbf{82.1}        & 94.8           & \textbf{92.2}           \\ \hline
2048              & 86.7       & 90.0        & 98.4           & 121.8          \\ \hline
\end{tabular}
\label{tab:ppl_position}
\end{table}

On both corpora, under the best settings, the TransfoRNN models showed lower PPLs indicating the redundancy of the positional encoding to the TransfoRNN models (as bolded in the Table \ref{tab:ppl_position}). Nevertheless, there are already Transformer LMs suggested that the positional encoding is not required for deep Transformer architecture \cite{Irie2019, Irie2020}.

\subsubsection{Inference in the Transformer models}

The Transformer models depicted in this work were evaluated by using the approach discussed in \cite{Al-Rfou2019}. The Transformers are suggested to use only the representation in the final position, i.e. $z_{i-1}^{(N)}$ for inference, instead of using the entire representations. Althoguh certain gain in terms of PPL might be obtained, the inference time would drastically increase, as proportion to the length of the considered context. 

In order to validate our results, we compared both settings in the Transformer models. Furthermore, as the approach was originally experimented at the character level \cite{Al-Rfou2019}, it's crucial to assess also at the word level, as how our models have been configured here. The results are shown in Table \ref{tab:ppl_transfp}.

\begin{table}[H]
\caption{PPLs of the Transformer models evaluated by using all or only the representation in the final position. }
\centering
\begin{tabular}{|l|c|c|c|c|c|}
\hline
\multirow{2}{*}{\textit{d}} & \multirow{2}{*}{\textit{N}} & \multicolumn{2}{c|}{PTB} & \multicolumn{2}{c|}{WikiText-2} \\ \cline{3-6} 
                   &                    & all         & final      & all            & final             \\ \hline
512                & 2                  & 107.2       & 103        & 133.4          & 124.3          \\ 
                   & 4                  & 96.4        & 94.0       & 116.9          & 108.5          \\ 
                   & 8                  & 94.3        & 88.5       & 106.3          & 98.6           \\ 
                   & 16                 & 102.9       & 94.4       & 105.9          & 96.9           \\ \hline
1024               & 2                  & 113.5       & 111.3      & 137            & 130.9          \\ 
                   & 4                  & 104.2       & 99.3       & 121.3          & 109.7          \\ 
                   & 8                  & 102.4       & 94.7       & 110.1          & 101.7          \\ 
                   & 16                 & 107.0       & 100.3      & 111.3          & 103.3          \\ \hline
\end{tabular}
\label{tab:ppl_transfp}
\end{table}

As shown in Table \ref{tab:ppl_transfp}, evaluating the Transformer models based only on the representation in the final position consistently shows lower PPLs, under all settings. Hence, we are assured that the proposed TransfoRNN models were compared with a more competitive Transformer model.

However, the TransfoRNN models discussed in this paper consistently used the entire representations for prediction, which we found out to give lower model PPLs.

\subsection{Speech recognition}

We also evaluated the performance of the TransfoRNN model on a speech recognition \textit{N}-best re-ranking task ny using the LibriSpeech corpus \cite{Panayotov2015}. The system was built followed the recipe in the Kaldi toolkit \cite{Povey2011}. The acoustic model comprises 17-layer TDNN taking 40-dimension MFCC and 100-dimension $i$-vector as input. The LM used in the decoder was the original official trigram model. 

Both TransfoRNN and Transformer models were trained from the transcripts in the training set, i.e. 960-hour, consisting of 9.4M words. The vocabulary contains 200K words. The TransfoRNN models consists of two Transformer layers followed by two LSTM layers while the counterpart Transformer models consists of two to eight layers. Other settings in both model are the same: eight attention heads, 1024-dimension embedding and 2048-dimension hidden layer.

The PPLs and WERs of the TransfoRNN and Transformer models were compared on the four datasets in the corpus: dev\_clean, test\_clean, dev\_other and test\_other. The results are shown in Table \ref{tab:libri_ppl} and \ref{tab:wer}.

\begin{table}[H]
\caption{PPLs of the TransfoRNN and Transformer models. The TransfoRNN model which comprises two Transformer layers slightly outperforms the Transformer model with eight layers.}
\centering
\begin{tabular}{l|c|cc|cc}
\hline
\multirow{2}{*}{} & \multirow{2}{*}{\textit{N}} & \multicolumn{2}{c}{clean} & \multicolumn{2}{c}{other} \\ 
                  &                    & dev         & test        & dev         & test        \\ \hline
Fourgram          & -                  & 283.7       & 289.5       & 250.8       & 259.2       \\ \hline
Transformer       & 2                  & 177.7       & 182.5       & 162.7       & 166.9       \\
                  & 4                  & 147.9       & 152.2       & 134.9       & 138.9       \\
                  & 8                  & 131.2       & 135.2       & 119.7       & 122.6       \\ \hline
TransfoRNN        & 2                  & \textbf{128.2}       & \textbf{133.3}       & \textbf{117.8}       & \textbf{119.8}       \\ \hline
\end{tabular}

\label{tab:libri_ppl}
\end{table}

\begin{table}[H]
\caption{WERs of the TransfoRNN and Transformer models. The results of both models are comparable.}
\centering
\begin{tabular}{l|c|cc|cc}
\hline
\multirow{2}{*}{} & \multirow{2}{*}{\textit{N}} & \multicolumn{2}{c|}{clean} & \multicolumn{2}{c}{other} \\ 
                  &                    & dev         & test        & dev         & test        \\ \hline
Trigram          & -                  & 3.41        &  3.85       & 9.32        & 9.35        \\ \hline
Transformer       & 2                  & 3.33        &   3.82      &   9.12      & 9.30        \\
                  & 4                  & 3.21        &    3.73     &  8.92       & 9.20         \\
                  & 8                  & \textbf{3.13}        &  3.72       &  \textbf{8.72}       & 9.09        \\ \hline
TransfoRNN        & 2                  & 3.19        &  \textbf{3.66}       &  8.80       & \textbf{9.02}        \\ \hline
\end{tabular}

\label{tab:wer}
\end{table}

For the PPLs, the TransfoRNN models which consists of only two Transformer layers consistently gave lower PPLs. These results are consistent to the results presented earlier. For the WERs, although both models reduced the baseline WERs, as output by the decoder, the TransfoRNN gave only comparable results in this evaluation.


\section{Conclusions}
\label{sec:conclusion}

This paper has discussed a work on using the recurrent layers to improve the Transformer LMs by encapsulating the sequential information into the models. The proposed model, referred to as TransfoRNN has shown lower model PPLs with fewer number of parameters in the models as compared to the vanilla Transformers. Moreover, we have shown empirically, that with the assistance of the recurrent layers, lesser number of Transformer layers is required. Specifically, a TransfoRNN model with only 2 layers of LSTM and 2 layers of Transformer outperformed a vanilla Transformer model with deeper network, e.g. 8 or 16. Through speech recognition evaluation on the Librispeech corpus, the TransfoRNN model showed comparable results with the Transformers.

For future work, the TransfoRNN shall be evaluated under BPE or character based settings with larger dataset, which is more applicable to the state-of-the-art end-to-end ASR systems.

\bibliographystyle{IEEEtran}

\bibliography{mybib}


\end{document}